# Audio Visual Emotion Recognition with Temporal Alignment and Perception Attention

**Linlin Chao[1], Jianhua Tao[1,2], Minghao Yang[1], Ya Li[1] and Zhengqi Wen[1]**
[1]National Laboratory of Pattern Recognition Institute of Automation, Chinese Academy of Sciences
[2]Institute of Neuroscience, State Key Laboratory of Neuroscience, CAS Center for Excellence in Brain Science and Intelligence Technology, Shanghai Institutes for Biological Sciences, CAS
{linlin.chao, jhtao, mhyang, yli, zqwen}@nlpr.ia.ac.cn

## Abstract

This paper focuses on two key problems for audio-visual emotion recognition in the video. One is the audio and visual streams temporal alignment for feature level fusion. The other one is locating and re-weighting the perception attentions in the whole audio-visual stream for better recognition. The Long Short Term Memory Recurrent Neural Network (LSTM-RNN) is employed as the main classification architecture. Firstly, soft attention mechanism aligns the audio and visual streams. Secondly, seven emotion embedding vectors, which are corresponding to each classification emotion type, are added to locate the perception attentions. The locating and re-weighting process is also based on the soft attention mechanism. The experiment results on EmotiW2015 dataset and the qualitative analysis show the efficiency of the proposed two techniques.

## 1 Introduction

Emotion recognition plays an important role in human machine interaction. Early researches mainly focus on utterance level speech emotion recognition or static image level facial expression recognition. However, emotion is a temporally dynamic event which can be better inferred from both audio and video feature sequences. This point of view is proved by cognitive researchers, where they argue that the dynamics of human behaviors are crucial for their interpretation [Sander et al., 2005]. Moreover, a number of recent studies [Chao et al., 2014; Liu et al., 2014] in affective computing demonstrate this point of view.

Meanwhile, human emotions are expressed in a multimodal way. Psychological study such as [Russell and Fernández-dols, 2003], has highlighted the importance of using multiple modalities to strengthen the accuracy of the emotion analysis. In [Busso et al. 2004], the authors analyzes the strengths and weaknesses of vision-only and audio-only based expression analysis systems. They also outline approaches for fusing the two modalities, and it is shown that when these two modalities are fused, the performance and the robustness of the emotion recognition system improve measurably.

Although combining audio and visual modalities improve the recognition accuracy, the audio visual fusion is still a problem. Three fusion strategies are widely utilized. Currently, most of the works combines the two modalities in decision level [Liu et al., 2014; Kahou et al., 2015]. In the decision-level fusion, the inputs coming from different modalities are modeled independently, and these single-modal recognition results are combined in the end. Since humans display audio and visual expressions in a complementary redundant manner, the assumption of conditional independence between audio and visual data streams in decision-level fusion is incorrect and results in the loss of information of mutual correlation between the two modalities [Zeng et al., 2005].

Feature level fusion is another way utilized in audio visual emotion recognition. [Sikka et al., 2013] combines visual descriptors with audio features using Multiple Kernel Learning and the audio-video clips are classified by SVM classifier. A more common way is extracting audio and visual features separately, pooling these features to single vectors for each feature sets and then concatenating these vectors into one single feature vector for classification [Busso et al., 2004; Chao et al., 2016]. These feature level fusion methods do not consider the temporal coupling of the audio and visual streams.

To address this problem, model-level fusion is proposed to make use of the correlation between audio and visual streams. In particular, multiple stream Hidden Markov Models (HMM) [Song et al., 2004; Zeng et al., 2005], Artificial Neural Network-based fusion [Fragopanagos and Taylor, 2005] are proposed. However, these models all require different modalities have strong synchronization. While the audio and visual signals always have different frame rates, temporal alignment before audio and visual features fed into these models is necessary, which is often manually operated.

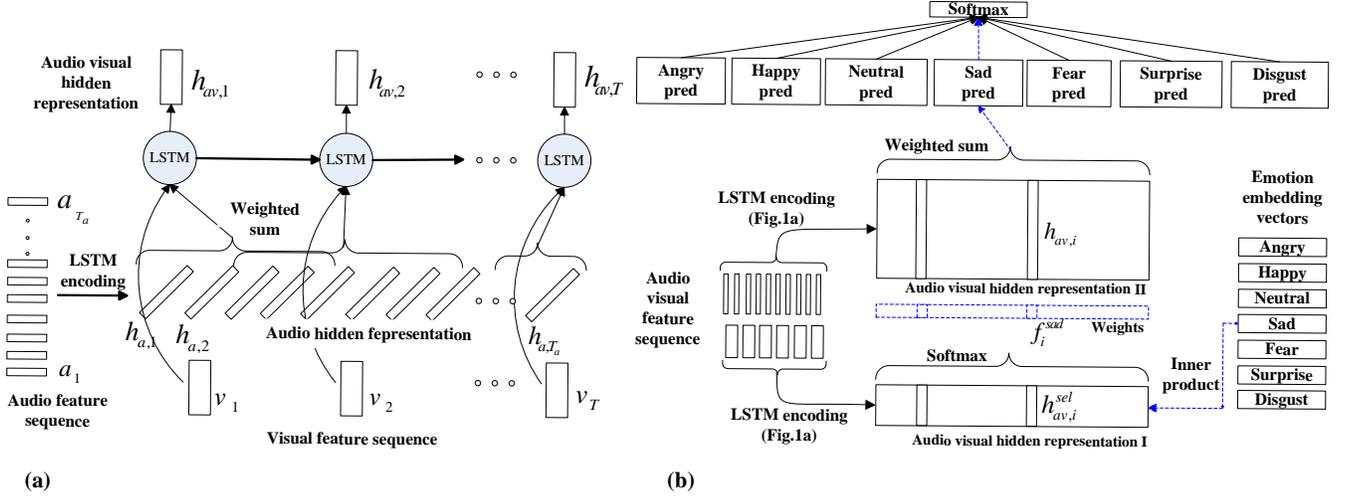

Figure 1: (1a): Soft attention aligns the visual feature sequence and audio representation sequence (encoded by a LSTM layer). The temporal aligned audio and visual streams are encoded by a LSTM layer to learn the dynamics and audio visual coupling (Section 2.1). (1b): After LSTM encoding, the added emotion embedding vectors locate and re-weight the perception attentions from the audio-visual stream (from Fig.1a) by soft attention. The final classification results are based on the combined weighted perception attentions (Section 2.2).

In this paper, we utilize LSTM-RNN [Hochreiter and Schmidhuber, 1997] to model the audio and visual streams. Particularly, soft attention mechanism [Mnih et al., 2014; Xu et al., 2015; Bahdanau et al., 2014] is employed for audio and visual streams alignment. This mechanism enables the neural network to learn to align audio and visual streams and predict emotion type jointly. Without manually temporal alignment, we believe this model can have less information loss and better recognition results.

When the RNN models are utilized for sequence classification, every time step outputs a hidden representation, which encodes the input information from start to the current time step. The final classification result is often calculated by the hidden reprensetation of the last time step (last-time encoding) [Kahou et al. 2015]. Previous study [Chao et al., 2016] shows that average all the hidden representations from different time steps (average encoding) can have better results. However, are the last-time encoding and average encoding the optimal choices? We believe that during the perception process for a special audio-video clip, people's attention will focus on several key sub-clips, which is more emotionally salient. These sub-clips can provide better clues and more attention should be paid for better emotion perception. Studies in the video emotion recognition filed can also prove this point of view. For example, [Kayaoglu and Eroglu Erdem, 2015] select the key frames in the video clip to make the final classification, which also shows competitive performance. The key frames can be seen as the emotional salient part.

Inspired by the above findings, we utilize the soft attention mechanism to re-weight and combine the hidden representations of all time steps in RNN for final classification. In order to locate the emotional salient parts, emotion embedding vectors, which are corresponding to each emotion classification types, are added to the proposed model. Each emotion embedding vector works as an anchor to choose and increase the weights of the salient parts of specific emotion type. After locating, re-weighting and combining, each emotion type will have a unique hidden representation for final classification. These emotion embedding vectors are jointly learned by the neural network with other parameters.

## 2 Method

### 2.1 LSTM model and audio visual alignment

We use the implementation discussed in [Xu et al., 2015]:

$$\begin{pmatrix} i_t \\ f_t \\ o_t \\ g_t \end{pmatrix} = \begin{pmatrix} \sigma \\ \sigma \\ \sigma \\ tanh \end{pmatrix} M \begin{pmatrix} h_{t-1} \\ x_t \end{pmatrix}, \quad (1)$$

$$c_t = f_t \odot c_{t-1} + i_t \odot g_t, \quad (2)$$

$$h_t = o_t \odot tanh(c_t), \quad (3)$$

where $i_t$ is the input gate, $f_t$ is the forget gate, $o_t$ is the output gate and $g_t$ is calculated by Eq.1. $c_t$ is the cell state, $h_t$ is the hidden state and $x_t$ represents the input to the LSTM at time step $t$. $M: R^a \to R^b$ is the affine transformation consisting of trainable parameters with $a = d + D$ and $b = 4d$, where $d$ is the dimensionality of all of $i_t$, $f_t$, $g_t$, $o_t$, $c_t$ and $h_t$ and $D$ is the dimensionality of the $x_t$.

In the proposed architecture, given the audio input feature sequence $a = \{a_1, a_2, ..., a_{T_a}\}$, a LSTM layer (Audio LSTM)

first learns the dynamics of the audio sequence and encodes it into hidden representation $h_a = \{h_{a,1}, h_{a,2}, ..., h_{a,T_a}\}$. The visual feature sequence is represented by $v = \{v_1, v_2, ..., v_T\}$. $T_a$ and $T$ represent the length of the audio feature sequence and visual feature sequence separately. The visual representation dynamics and audio visual coupling are encoded together by another LSTM layer (Audio-Visual LSTM). In this layer, soft attention mechanism is utilized to align the audio and visual streams. During alignment, window technique is applied. At each time step $t$, the soft attention mechanism considers a sub-sequence $h_{att} = \{h_{a,p_t-w}, ..., h_{a,p_t+w}\}$ of the whole sequence $h_a$, where $w$ is a predefined window width and $p_t$ is the median of the alignment. Given $T_a$ and $T$, we can calculate the coarse alignment of the two streams, which is $p_t$ for each time step. Adding a window can utilize this prior knowledge and also result in a lower complexity. The more accurate alignment $l_t$ can be calculated as the follow ways:

$$l_{t,i} = \frac{\exp(s_{t,i})}{\sum_{j=1}^{2*w} \exp(s_{t,j})} \quad i \in 1, ..., 2*w, \quad (4)$$

$$s_{t,i} = W_i^T \tanh(W^a h_{att,i} + W^v v_t), \quad (5)$$

where $W_i$ are the weights mapping to the $i^{th}$ element of the softmax, $s_{t,i}$ is the score of $h_{att,i}$ aligned to $v_t$ and these scores are normalized by softmax (Eq. 4). The softmax can be thought of as the probability with which our model believes the corresponding frames in the $h_{att}$ are temporal aligned to $v_t$. After calculating these probabilities, the soft attention mechanism computes the expected value of the $h_a$ at every time step $x_t$ by taking the expectation over $h_{att}$:

$$x_t = \sum_{i=1}^{2*w} l_{t,i} h_{att,i}. \quad (6)$$

Then $x_t$ and the corresponding $v_t$ are fed into the audio-visual LSTM, which learns the correlation and dynamic of the audio and visual streams (see Fig.1a.).

At every time step $t$, hidden representation $h_{av,t}$ of the audio-visual LSTM is calculated, which encodes the input features from start to time step $t$. Normally, there are two ways to get the final classification results. The first one (last-time encoding) is based on $h_{av,T}$, which is the last hidden representation. The classification results via last time encoding can be represented as:

$$p(y|(v, x)) = p(y|(v_1, x_1), (v_2, x_2), ..., (v_T, x_T)). \quad (7)$$

The other one is the average encoding, which is calculated based on the average of the whole $h_{av}$. Classification results via average-pool can also be represented as:

$$p(y|(v, x)) = \sum_{t=1}^{T} p(y|(v_1, x_1), (v_2, x_2), ..., (v_t, x_t)). \quad (8)$$

For the last-time encoding, the $h_{av,T}$, a fixed-length vector, may not fully contain all the necessary information from the whole audio-visual stream. This problem also exists in RNN based machine translation [Bahdanau et al., 2014]. Previous study [Chao et al., 2016] has proved the average encoding is better than the last time encoding. However, average encoding encodes $h_{av}$ as a classifier fusion way with equal weights given to the sub-parts of the whole sequence. As perception attention exists when perceiving the audio-visual clips, given equal weights to each sub-parts is not an optimal way. Thus, better encoding way should be explored. Find the perception attentions and increase the weights of perception attentions is a solution.

### 2.3 Perception attention

We hypothesis different emotion types have different perception attentions. To locate these perception attentions, we add emotion embedding vectors $e = \{e_1, e_2, ..., e_N\}$ to the model. $N$ equals to the number of emotion types which needs to be classified. Each emotion embedding vector works as an anchor to select the emotional salient parts from the whole audio-visual stream. Based on the soft attention mechanism, the attention scores $f^n = \{f_1^n, f_2^n, ..., f_T^n\}$ for emotion type $n$ are calculated as follows:

$$f_i^n = \frac{\exp((W^h h_{av,i})^T e_n)}{\sum_{j=1}^{T} \exp((W^h h_{av,j})^T e_n)} \quad i \in 1, ..., T, \quad (9)$$

where $W^h$ is the mapping matrix for $h_{av}$ from the orignal dimension to the dimension of $e_n$. $f^n$ represents emotion type $n$'s attention distribution of $h_{av}$. Then we can get the hidden representation $\sum_{i=1}^{T} f_i^n h_{av,i}$, which is specially calculated for emotion type $n$. The classification score $s^n$, which is the audio-visual clip is classified to emotion type $n$, is calculated as:

$$s^n = W^n \sum_{i=1}^{T} f_i^n h_{av,i} + b^n, \quad (10)$$

where $W^n$ and $b^n$ are assigned to emotion type $n$, and $W^n$: $R^a \to R^b$ with $a$ equals to the dimension of $h_{av,i}$ and $b$ equals to one. The classification scores of all the emotion types are then normalized by softmax function,

$$p(y = n|(a, v)) = \frac{\exp(s^n)}{\sum_{j=1}^{N} \exp(s^j)} \quad n \in 1, ..., N. \quad (11)$$

where $y$ represents the predicted emotion type.

By adding embedding vectors, the model can learn the perception attentions for each emotion type. Then output the classification score based on the weighted combination of perception attentions. Compared to average encoding or last-time encoding, the hidden representations of LSTM-RNN are utilized more efficiently and the information loss is decreased.

In the implementation of the proposed model, we encode the audio and visual streams two times with similar architecture. The first time encodes $(a, v)$ to hidden representation $h_{av}^{sel}$, which are utilized to locate the attentions and obtain $f^n$ in Eq.9, with a smaller network. The second time encodes $(a, v)$ to $h_{av}$. Then $f^n$ and $h_{av}$ combines for further processing (see Fig.1b). When we calculate $f^n$ and $h_{av}$ separately, better performance can be obtained compared to compute $f^n$ based on the same $h_{av}$. We believe the separate encoding way can decrease the correlation among parameters. Thus it is easier to optimize.

## 3 Dataset and Feature Set

The EmotiW2015 [Dhall et al., 2015] provides the common benchmarks for emotion recognition researchers, which mimics real-world conditions. There are two sub-challenges: audio-visual based emotion recognition challenge (AFEW) and image based static facial expression recognition challenge (SFEW). AFEW sub-challenge is to assign a single emotion label to the video clip from the six universal emotions (Anger, Disgust, Fear, Happiness, Sadness and Surprise) and Neutral. The databases (AFEW and SFEW) are divided into three sets for the challenge: training, validation and testing. The training and validation sets are utilized to train the emotion recognizer. Prediction results on testing set are utilized to rank participants. The sample rate for audio data in AFEW is 44kHz. The video data in AFEW has 25 frames per second.

### 3.1 Face shape feature

For video features, we mainly focus on the face part. As the face shape provides import clues for facial expression, we use the landmarks' location of the face as face shape feature. After feature normalization for each clip, these features can also reflect the head movement and head pose. The 49 landmarks' locations are then PCA whitened [Bengio, 2012], with the final 20 dimensions are kept.

### 3.2 Face appearance feature

For face appearance feature, we utilize the features extracted from a convolutional neural network (CNN) [LeCun et al., 1998] model. Previous work [Liu et al. 2014] utilizes the CNN model trained via face recognition dataset to extract face representation. And this representation can be generalized to facial expression recognition problem. We employ the same strategy to train a CNN model from Celebrity Faces in the Wild (CFW) [Zhang et al., 2012] and Facescurb [Ng and Winkle, 2014] dataset, which are designed for face recognition tasks. Over 110,000 face images from 1032 people are used for training and the labels are their identities. The architecture is the same with [Krizhevsky et al., 2012]. There are three fully connected layers and five convolutional layers. Compared to the three fully connected layers, convolutional layers have better generation performance [Girshick et al., 2013]. The deeper layers extract more abstract features [Zeiler and Fergus, 2013]. Thus, we extract the feature from the 5th pooling layer (pool5) as appearance feature. While the dimension number of the features from pool5 is 9216. Meanwhile, the training data is relatively small. Thus we employ random forest algorithm implemented by scikit-learn[1] for feature selection and 1024 features are kept for the appearance feature set. The random forest classifier is trained via the SFEW database, where one of the seven emotion labels is assigned to a single static face image.

### 3.3 Audio feature

We utilize the YAAFE toolbox[2] to extract audio features. All the 27 features of the toolbox are extracted. The audio data is resampled to 16KHz and default parameters of each feature is utilized. Finally, 939 dimensions features are extracted for each frame and the frame length is 1024. The audio features are then PCA whitened, with the final 50 dimensions are kept.

## 4 Experiments

### 4.1 Experiments setup

We follow the challenge criterion of EmotiW2015 to utilize training set, validation set and testing set. We utilize the landmarks provided by the organizers for the shape feature. Caffe [Jia et al., 2015] implementation of CNN is utilized to extract face appearance features, where the cropped face image is provided by the organizers.

For the verification of perception alignment, we compare the average encoding or last-time encoding with the proposed model. Thus, there are mainly two architectures for comparison. The first one is the average encoding or last-time encoding for the audio visual inputs. In this architecture, there are 64 memory cells utilized for both Audio LSTM and Audio-Visual LSTM. The dimensions of all the hidden layers before LSTM layers are equal to the dimension of LSTM layers. The audio feature sequence is fed into a hidden layer first and then fed into the Audio LSTM. For visual feature sequences, both face shape feature and face appearance feature are fed into one hidden layer with 64 nodes separately. The hidden representations of the two feature sets are then concatenated together and fed into another hidden layer. The fused hidden representation of face shape and appearance features is utilized to align the audio stream and represent the visual representation for Audio-Visual LSTM. The second one is the proposed model with both temporal alignment module and perception attention module. The main difference compared with the first architecture exists in the perception attention module (Section 2.2) is added. The same architecture with the first one is utilized to get $h_{av}^{sel}$ and $h_{av}$, with the dimensions are 32 and 8. The dimension of the seven emotion embedding vectors is also 8. Thus, the second model have smaller parameters (In the experiments, we find the model works best when the memory cell number of LSTMs for $h_{av}$ equals to 32).

---

[1] http://scikit-learn.org/stable/
[2] http://yaafe.sourceforge.net/

For the verification of audio visual alignment, the performance of the average encoding with single face appearance feature and the audio visual inputs with temporal alignment module only are compared. The model for appearance feature has one hidden layer and one LSTM layer with 64 memory cells.

All the models are trained via Adadelta [Zeiler, 2012]. The maximum training epoch is 50 with dropout regularization technique utilized in all layers except the LSTM layer. The drop rate is 0.5. Weight decay in all the layers with the parameter 0.0005 is applied to prevent over fitting. Early stopping technique is also employed. The best results for testing set are chosen by the best prediction accuracy in the validation set.

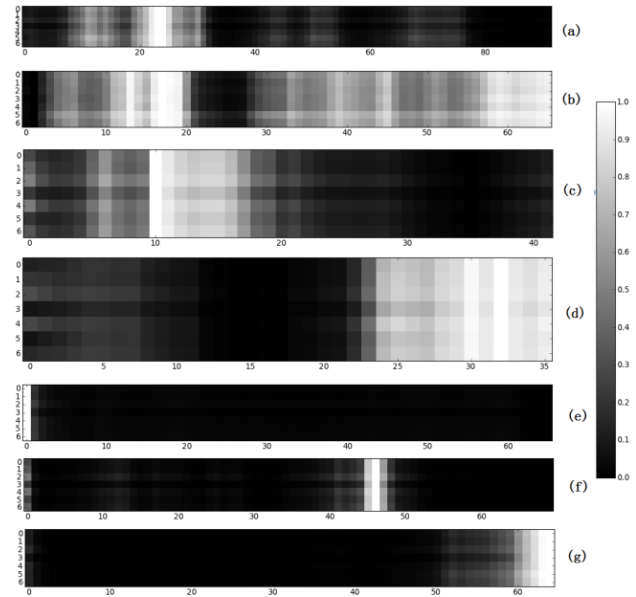

Figure 3: Examples of perception attentions for each emotion type to the audio visual stream. The gray values of the bars indicate the attention scores. Each row is normalized to have maximum value of 1. The rows of each sample are corresponding to the seven emotion types, which are angry, disgust, fear, happy, neutral, sad and surprise. The columns are corresponding to the time steps of each sample.

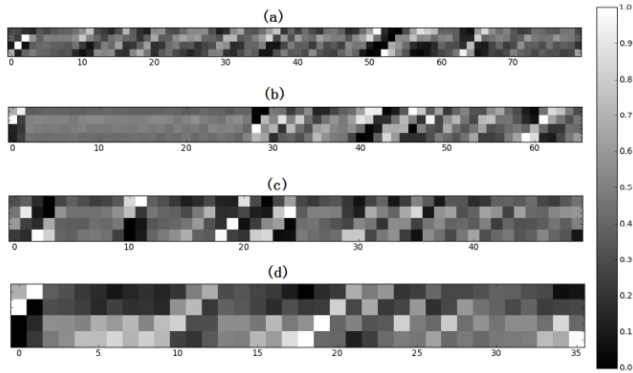

Figure 2: Examples of temporal alignment for audio and visual streams. The gray values of the bars indicate the scores of the alignment from the audio frames in an given window to a particular visual frame. Each row is normalized to have maximum value of 1. The rows of each sample are corresponding to the four locations in the given window and the columns are corresponding to the time steps. The bottom row of each sample is the first location in the given window.

### 4.2 Qualitative analysis

In Fig.2, we show how the alignments between audio and visual streams change with the increase of time step. Looking at the overall, the examples show a clear shift of the attention focuses in the given window as time step increases (shift at 45 degree visually and shift from one location in an window to next location in next window). Between the obvious attention shift time steps, the middle time steps show no clear change mode. This is because as time step increases, the attentions will shift from start to end of the window and shift reverse from end to the start happens in these middle time steps. The changing mode happens in almost all samples no matter how many time steps of each sample (Fig. 4a-4d). The first half of Fig. 4b also shows different attention distribution, which is similar to even distribution. The reason for this is mainly from the dataset. As the EmotiW dataset is collected from the movies, which are in a wild environment and the background can be very noise [Dhall et al., 2015]. When the audio modality has no sound of human, the alignment tends to become even distribution.

The perception attention visualization shows in Fig.3. As the perception attentions have various distributions for different samples, we randomly pick several samples to represent all the samples. This figure shows that perception attentions can locate different parts in the whole audio-visual streams, from the start (Fig.3e), close to middle (Fig.3f) to the end (Fig.3g). This figure also shows that the perception can locate in multiple locations (mainly in the first half (Fig.3a and Fig.3c), in the second half (Fig. 3d) and random distribution in the whole stream (Fig.3b)). In the picking process, we also find that a relative large proportion of samples have the same distribution with Fig.3e, where the perception attention mainly locates in the first frames. Two reasons may explain this result. The first one is that the neural network fails to learn the right distribution totally. The second one may come from the dataset collection process. During the annotators label the audio-visual recording, the clip begins when they find the emotional salient part. Thus, the start of the audio-visual clip can be the most emotional salient frame. In this context, seven emotion embedding vectors are added. We can see the seven emotion type almost focus on the same emotional salient sub-parts with relative little differences in the attention distribution. This suggests that the emotional salient parts attract our attention to judge the emotion types.

Fig.4 shows the projection of the emotion embedding vectors from random initialization before model training (Fig.4a) to the final values when the model training finished (Fig.4b). The relative positions of these vectors change totally. However, there are not clear patterns among the relative positions of each vector when jointly observing with the confusion matrix (Fig.5).

More details of the best submitted result are shown in Fig.5. The confusion matrix shows that angry, happy, neutral and sad are easier to classify. The surprise and fear are easy to be misclassified to angry. Disgust is easy to confuse with happy. The reason may lie in the data set distribution is not balance. Fine grained classification among angry, fear and surprise needs more effort.

| | Angry | Disgust | Fear | Happy | Neutral | Sad | Surprise |
|---|---|---|---|---|---|---|---|
| Angry | **75.95** | 0 | 8.86 | 1.27 | 5.06 | 8.86 | 0 |
| Disgust | 3.45 | 6.90 | 1.03 | **20.69** | 31.03 | 24.14 | 3.45 |
| Fear | **36.36** | 4.54 | 22.73 | 0 | 13.64 | 7.58 | 15.15 |
| Happy | 12.84 | 1.83 | 0.92 | **55.96** | 7.33 | 19.27 | 0.92 |
| Neutral | 7.55 | 4.40 | 8.18 | 8.18 | **47.80** | 18.24 | 5.66 |
| Sad | 11.11 | 1.39 | 12.5 | 9.72 | 19.44 | **33.33** | 11.11 |
| Surprise | **25.93** | 0 | 18.52 | 7.40 | 14.81 | 18.52 | 14.81 |

Figure 5: Confusion matrix on the testing set of the proposed model.

| Model | Accuracy % | | |
|---|---|---|---|
| | Train | Val | Test |
| average encoding (pool5 only) | 73.84 | 43.14 | 39.89 |
| last-time encoding | 82.70 | 36.39 | |
| average encoding | 83.83 | 43.40 | 41.19 |
| proposed model | 58.23 | 46.90 | 44.90 |

Table 1: Experiment results on training, validation and testing set for Audio-Video Emotion Recognition sub-challenge.

| Method | Acc % |
|---|---|
| AU-aware features + SVM (Yao et al. 2015) | 53.8 |
| Spatial-temporal features+PLS+ELM (Kaya et al., 2015) | 53.6 |
| CNN-RNN + Decision Fusion (Kahou et al., 2015) | 52.9 |
| Our model | 44.9 |

Table 2: Comparison of performance on for Audio-Visual Emotion Recognition sub-challenge with state-of-the-art models.

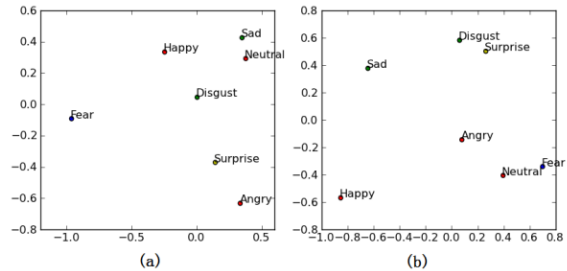

Figure 4: The projection of the emotion embedding vectors. All the vectors are projected to a 2-D space via PCA dimension reduction and the top two eigenvalues are kept for each vector. (4a): the projection of the initialized embedding vectors before the model training. (4b): the projection of the learned embedding vectors.

### 4.3 Quantitative analysis

Table 1 reports accuracy of the comparison experiments. The results show that the performance is slightly better when combine the audio visual modalities in feature level. The proposed model also works better than the average encoding and last-time encoding model.

Table 2 also shows the performance comparisons with several state-of-the-art models. The first three results are the top three performers on EmotiW 2015. There are significant gaps compared with these leading models. The top 2 performers all focus in designing better features. The work of [Kahou et al., 2015] is more close to our work. However, their model utilizes decision level fusion to improve the performance significantly. In fact, their RNN model with appearance feature behaves similar to ours. We can conclude that decision level model works better than feature level fusion on this dataset and better features are needed for our model.

## 5 Conclusion

In this paper we utilize the soft attention mechanism to temporally align the audio and visual streams and fuse these streams in the feature level. We also add the emotion embedding vectors and the soft attention mechanism in the output layer of RNN to locate and re-weight the perception attentions in the audio visual stream. Compared to the widely utilized average encoding or last-time encoding, our model decrease the information loss in the output layer of RNN and utilize the output of RNN more efficiently. Besides, both the qualitative analysis and quantitative analysis show the effectiveness of the proposed techniques. We also think the proposed model, especially for the perception attention technique, can be utilized to other sequence classification tasks. In the future, we plan to explore better features for emotion classification task since there is still large space to improve compared with the state-of-the-art models.

## Acknowledgments

This work is supported by the National High-Tech Research and Development Program of China (863 Program) (No.2015AA016305), the National Natural Science Foundation of China (NSFC) (No.61425017, No.61332017, No.61375027, No.61305003, No.61203258), the Major Program for the National Social Science Fund of China (13&ZD189) and the Strategic Priority Research Program of the CAS (Grant XDB02080006).